\documentclass{bvm} 

\makeatletter
\let\blx@rerun@biber\relax
\makeatother

\begin{document}



\selectlanguage{english} 


\title{Taming Detection Transformers for Medical Object Detection}



\titlerunning{DETR for Medical Object Detection}


\author{
    Marc~K. \lname{Ickler}\inst{* 1}, 
    Michael \lname{Baumgartner}\inst{* 1, 3, 4},
    Saikat \lname{Roy} \inst{1, 3},
    Tassilo \lname{Wald} \inst{1, 4},
    Klaus~H. \lname{Maier-Hein} \inst{1,2}
}

\authorrunning{Ickler et al.}

\institute{\inst{1} Division of Medical Image Computing, German Cancer Research Center, Heidelberg, Germany\\
\inst{2} Pattern Analysis and Learning Group, Heidelberg University Hospital\\
\inst{3} Faculty of Mathematics and Computer Science, Heidelberg University, Germany\\
\inst{4} Helmholtz Imaging\\
\inst{*} equal contribution\\ 
}

\email{m.baumgartner@dkfz.de}

\maketitle

\begin{abstract}
The accurate detection of suspicious regions in medical images is an error-prone and time-consuming process required by many routinely performed diagnostic procedures.
To support clinicians during this difficult task, several automated solutions were proposed relying on complex methods with many hyperparameters.
In this study, we investigate the feasibility of DEtection TRansformer (DETR) models for volumetric medical object detection.
In contrast to previous works, these models directly predict a set of objects without relying on the design of anchors or manual heuristics such as non-maximum-suppression to detect objects. 
We show by conducting extensive experiments with three models, namely DETR, Conditional DETR, and DINO DETR on four data sets (CADA, RibFrac, KiTS19, and LIDC) that these set prediction models can perform on par with or even better than currently existing methods. DINO DETR, the best-performing model in our experiments demonstrates this by outperforming a strong anchor-based one-stage detector, Retina U-Net, on three out of four data sets.

\end{abstract}

\section{Introduction}
Clinical decision-making is often based on the correct localization and classification of multiple pathologies throughout the entire human body.
Because this task is frequently associated with an error-prone and time-sensitive diagnostic process, multiple computer-aided diagnostic (CAD) systems have been published in the past to automate this process.
Most of the previous works relied on methods with anchors~\cite{3350-Baumgartner2021, 3350-Jaeger2020} or center points suffering from complex design and manual heuristics, such as the anchor matching strategy or non-maximum suppression.
Recently, a new detection paradigm was proposed~\cite{3350-Carion2020} for natural images by reformulating the detection task as a set prediction problem. In contrast to previous methods, these models directly predict a set of objects without relying on manual heuristics such as anchors and post-processing procedures.
While the first introduction of the \emph{DEtection TRansformer} (DETR)~\cite{3350-Carion2020} already demonstrated impressive performance, it suffered from reduced performance on small objects and long training times. Follow-up work~\cite{3350-Meng2021, 3350-Zhang2022} tackled these shortcomings and is now achieving state-of-the-art (SOTA) results on the commonly used COCO benchmark.

Despite their advantages, detection transformers remain poorly studied in the medical domain. Only one study~\cite{3350-Wittmann2022} employed them to detect organs on three-dimensional CT images but the basic DETR models did not show promising performance compared to an anchor-based method. More research is required to assess the potential of detection transformers in the medical field because of the restricted experiments on one task conducted thus far. In this work, we aim to study the feasibility of DETR models on a diverse set of volumetric medical object detection problems. We evaluate three set prediction models, namely DETR~\cite{3350-Carion2020}, Conditional DETR~\cite{3350-Meng2021} and DINO-DETR~\cite{3350-Zhang2022} against the strong anchor-based baseline nnDetection~\cite{3350-Baumgartner2021} on four data sets (CADA \cite{3350-Ivantsits2022}, RibFrac \cite{3350-Jin2020}, KiTS19 \cite{3350-Heller2019} and LIDC \cite{3350-Armato2011}).

\section{Materials and methods}
\subsection{Network architecture}
\begin{figure}[t]
	\centering
	\setlength{\figwidth}{\textwidth}
 	\caption{A convolutional neural network (CNN) takes a patch and computes a feature map that is flattened to a sequence. After adding positional encoding, the sequence is fed through a transformer encoder and decoder. The output sequence is processed by two feed-forward networks (FFN), one predicting bounding boxes and the other class labels. Conditional DETR (green) separates content and spatial queries and uses conditional cross-attention to attend to important regions. The DINO DETR module's multi-scale features, anchors, and denoising queries (DN Q) are shown in red.}
	\includegraphics[width=\textwidth]{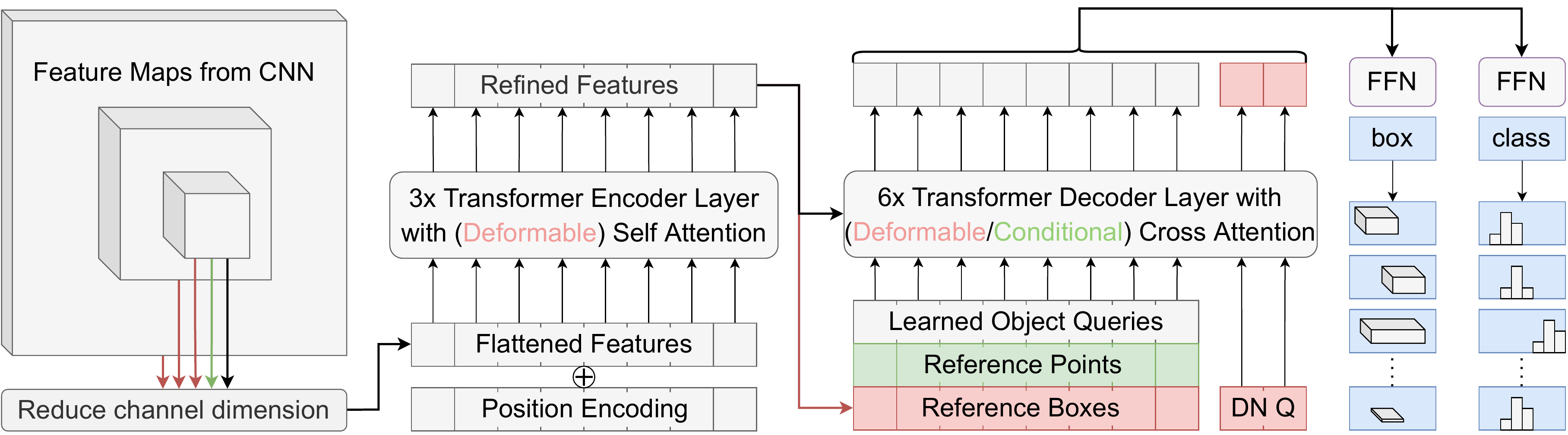}

	\label{3350-detr_arch}
\end{figure}

The architecture of DETR, Conditional DETR, and DINO DETR is shown in Fig.~\ref{3350-detr_arch}.
DETR \cite{3350-Carion2020} has two special properties that differentiate it from other object detectors.
Firstly, it uses a transformer, which takes in a sequence of features and outputs a set of object proposals. Owing to the global attention mechanism, the transformer is expected to detect large features in the encoder as well as prevent duplicate predictions in the decoder.
Secondly, DETR predicts bounding boxes with direct set prediction. This requires a loss that is invariant to permutations of the predictions. The Hungarian algorithm finds an optimal bipartite matching between the predictions and ground truth objects that is used to calculate the classification and regression losses. By using direct set prediction, no additional post-processing steps are required, but the model has to prevent duplicate predictions. DETR was the first method that does this in a non-autoregressive way by using a transformer with parallel decoding while achieving competitive performance to current anchor-based models. Because of the attention mechanism, the model can globally reason about all objects using the pair-wise relations between queries. This novelty makes the difficult and potentially performance-degrading non-maximum-suppression obsolete.

However, the vanilla DETR architecture converges slowly and struggles to detect small objects. Conditional DETR \cite{3350-Meng2021} introduces conditional attention and reference points to improve upon both weaknesses. In the decoder, each query predicts one reference point around which it will attend to encoder features and predict the output bounding box. Additionally, the architecture increases the number of queries and uses the focal loss instead of the cross-entropy loss to improve the classification. DINO DETR \cite{3350-Zhang2022} utilizes a sparse attention called deformable attention, which reduces memory consumption and enables the use of multi-scale features in the transformer. In addition, it takes features from the encoder to initialize the reference boxes. These boxes are not predefined unlike anchors in other methods and are used throughout the decoder to attend to encoder features and predict bounding boxes close to them. Additional denoising queries and denoising losses are implemented to improve the training signal for faster convergence. Finally, DINO DETR also uses focal loss and even more queries than Conditional DETR.

\subsection{Data sets}
To cover a diverse set of problems we performed experiments on four medical object detection data sets.
The CADA data set \cite{3350-Ivantsits2022} has 127 objects in 109 scans and covers the low data regime. The target structures are brain aneurysms which are typically small.
In the RibFrac data set \cite{3350-Jin2020} the task is to detect rib fractures. It represents a large medical data set with 4422 objects in 500 thin-slice CT scans.
For multi-class detection, we use the LIDC data set \cite{3350-Armato2011} and follow the previous design of \cite{3350-Jaeger2020} to divide lesions into two categories of lung nodules that are difficult to distinguish. It comprises of 1035 CT scans where the annotated lung nodules are very small making the task even more challenging. Finally, DETRs performance on large objects is studied on the KiTS19 data set \cite{3350-Heller2019} that has 225 objects in 204 cases. The target structures are kidney tumors.

\subsection{Experimental setup}
We use a patch-wise prediction-based approach and employ nnDetection \cite{3350-Baumgartner2021} as our development framework for training, inference and evaluation. Within the nnDetection augmentation pipeline, we decrease the maximum rotation to 20 degrees and increase the minimum downscaling to 0.8 to increase training speeds. For a fair comparison to Retina U-Net, all models are configured to require less than $11 \ts \text{GB}$ of GPU memory. We utilize the standard nnDetection backbone (a plain convolutional encoder) and use the configured, data set-dependent patch size. For DINO DETR, the three feature maps with the lowest resolutions are fed into the transformer. We change the hidden dimension to 120 or 128 for DINO DETR and the transformer feed-forward network dimension to 1024 to adapt the model to medical data. We use a different number of queries depending on the data set and architecture (Tab. \ref{3350-settings}).
\begin{table}[t]
    \caption{Data set specific settings for all data sets. Abbreviations: DE=DETR, CD=Conditional DETR, DI=DINO DETR. Only DINO DETR uses denoising queries.}
	\begin{tabular*}{\textwidth}{l@{\extracolsep\fill}lllllllllllll}
		\hline
		& \multicolumn{3}{l}{CADA} & \multicolumn{3}{l}{RibFrac} & \multicolumn{3}{l}{KiTS19} & \multicolumn{3}{l}{LIDC} \\
		\hline
        architecture& DE   & CD & DI    & DE    & CD    & DI    & DE    & CD & DI & DE & CD & DI\\
        \# queries (denoising)  & 6    & 12 & 20 (8)  & 20    & 50    & 90 (20)  & 6    & 12 & 20 (8) & 12  & 24  & 40 (8)\\

        \# epochs & 50 & 50 & 25 & 125 & 100 & 75 & 100 & 100 & 35 & 200 & 100 & 75 \\
        \hline
	\end{tabular*}
	\label{3350-settings}
\end{table}

The training schedule is PolyLR with an exponent of 0.9 and a learning rate of 0.0001. The AdamW optimizer with a weight decay of 0.0001 is used. We train for 2500 batches per epoch with a batch size of 4 and adapt the number of epochs for all models to converge (Tab. \ref{3350-settings}). Other parameters were left to their default values since we found them to translate well to the medical domain.

\section{Results}
\begin{figure}[b]
	\centering
	\includegraphics[width=\textwidth]{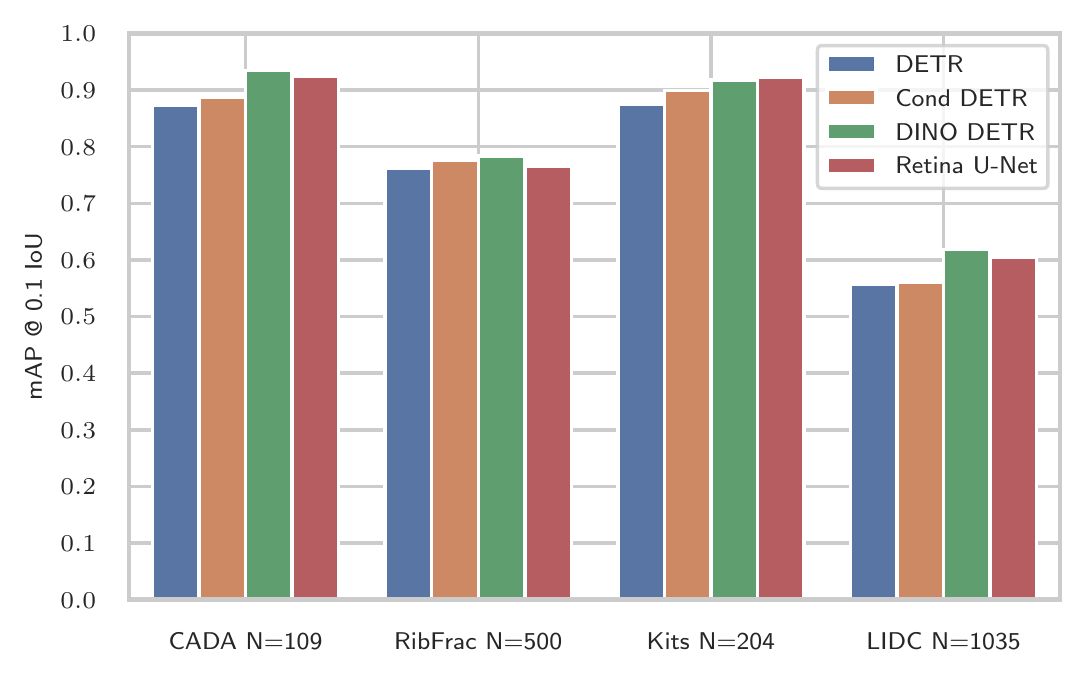}
	\caption{Performance comparison between Retina U-Net (red), DETR (blue), Conditional DETR (orange), DINO DETR (green) on CADA, RibFrac, KiTS19, and LIDC. $N$ denotes the number of scans in the data set.}
	\label{3350-performance_comparison}
\end{figure}

The mean average precision (mAP) metric at an Intersection-over-Union (IoU) value of 0.1 is used for evaluation to reflect the clinical need for coarse localizations. Reported results were obtained by performing a five-fold cross-validation on each data set.
The results are shown in Fig.~\ref{3350-performance_comparison}.

All models show similar performance with a maximum discrepancy smaller than ten percent per data set. Especially on the largest data set RibFrac, the model scores are close with DETR being the worst and DINO DETR being the best model. We observe a trend clearly showing that DINO DETR performs best, followed by Conditional DETR and DETR, in agreement with our initial architectural analysis. DINO DETR and Retina U-Net achieve remarkably similar scores, with DINO DETR performing slightly better on three out of four data sets. 

\section{Discussion}
Our results in Fig.~\ref{3350-performance_comparison} illustrate that detection transformers are a viable option for medical object detection. While not on par with Retina U-Net, even the vanilla DETR model with a few modifications for volumetric input shows promising performance. Further optimized models like Conditional DETR and DINO DETR close the gap and show that competitive scores can be achieved using set prediction models.

By omitting the need for anchors and non-maximum suppression, set prediction models depend on fewer hyperparameters and therefore reduce the search space during the hyperparameter tuning process. The number of predictions is controlled by the number of queries in the transformer decoder and is one of our only data set-dependent parameters. This simplicity makes detection transformers easy to adapt to different detection problems and great candidates for self-configuring models.

Furthermore, the detection transformers only need bounding box inputs for training. Retina U-Net uses semantic segmentation as an auxiliary task to improve detection performance, making it dependent on segmentation labels which take more effort to annotate. Our experiments show that DINO DETR can achieve similar results without using an auxiliary segmentation loss.

While the performance of the models is similar, their convergence speed is not. DETR takes 200 epochs to converge on LIDC, four times more than Retina U-Net. In general, DETR and Conditional DETR need more training than Retina U-Net and DINO DETR, but the detection transformer models have the advantage of being computationally less expensive. The transformer is small compared to the backbone and forward and backward passes are approximately $1.5$ times faster than for Retina U-Net (measured on a system equipped with a Ryzen 9 3900X, 64GB RAM, and an NVIDIA RTX 3090 24GB).

The most significant performance difference between DINO DETR and the other DETR models occurs on the CADA and LIDC data sets, which contain the smallest objects. We assume that the drop in performance stems from a lack of high-resolution features in the single-scale DETR models. DINO DETR uses three feature maps with the highest resolution being four times higher per dimension than the resolution of the feature maps used by DETR and Conditional DETR. This higher feature resolution improves the localization of objects in the transformer and increases performance. The resolution of DETR and Conditional DETR can be improved by decreasing the stride in the backbone and reducing the score gap to DINO, only requiring more GPU memory. To provide a fair comparison, this experiment was omitted here since we restricted the available GPU memory to $11 \ts \text{GB}$ while using the same patch size as the baseline model.

Additionally, the improved performance of DINO DETR comes at the cost of model complexity. While DETR is a straightforward model which can be easily implemented and understood, DINO DETR introduces more complex features and changes to most modules of DETR. Conditional DETR only introduces slight changes to the transformer decoder of DETR, keeping it more similar.

While this study establishes the competitive nature of the class of transformer-based set prediction methods (DETR) for object detection in medical images, there also exist some limitations. Firstly, the detection transformer models used are not self-configuring, and thus per data set tuning was required. Another limitation is that because of the patch-wise training and prediction, a strategy to merge predictions from overlapping patches has to be implemented. This reintroduces the need for some post-processing, but because of far fewer predictions for detection transformers, other methods could be applied.

Considering the promising results, a future study of a self-configuring detection transformer on all ten data sets proposed by nnDetection would be of high interest. The detection transformers can also efficiently be extended to instance segmentation models.

In conclusion, detection transformers are well-suited for medical object detection. They prove to have competitive performance on a wide range of data sets without the need for predefined anchors, non-maximum-suppression, or segmentation losses.

\begin{acknowledgement}
Part of this work was funded by Helmholtz Imaging, a platform of the Helmholtz Incubator on Information and Data Science.
\end{acknowledgement}
\printbibliography
\end{document}